\pdfoutput=1

\documentclass[11pt]{article}

\usepackage[]{naacl2021}

\usepackage{times}
\usepackage{latexsym}

\usepackage{multirow}
\usepackage{amsmath}
\usepackage{amssymb}
\usepackage{arydshln}
\usepackage[ruled,vlined,linesnumbered]{algorithm2e}
\usepackage{algpseudocode}
\usepackage{xcolor}
\usepackage{graphicx}

\usepackage[T1]{fontenc}

\usepackage[utf8]{inputenc}

\usepackage{microtype}

\title{Neural Sequence Segmentation as Determining the Leftmost Segments}

\author{Yangming Li\textsuperscript{\rm 1}, Lemao Liu\textsuperscript{\rm 1}, Kaisheng Yao\textsuperscript{\rm 2} \\
	\textsuperscript{\rm 1}Tencent AI Lab \\
	\textsuperscript{\rm 2}Ant Group, Alibaba Group \\
	\texttt{\{newmanli,redmondliu\}@tencent.com} \\
	\texttt{kaisheng.yao@antgroup.com} \\}

\begin{document}
\maketitle
	
\begin{abstract}

	Prior methods to text segmentation are mostly at token level. Despite the adequacy, this nature limits their full potential to capture the long-term dependencies among segments. In this work, we propose a novel framework that incrementally segments natural language sentences at segment level. For every step in segmentation, it recognizes the leftmost segment of the remaining sequence. Implementations involve LSTM-minus technique to construct the phrase representations and recurrent neural networks (RNN) to model the iterations of determining the leftmost segments. We have conducted extensive experiments on syntactic chunking and Chinese part-of-speech (POS) tagging across 3 datasets, demonstrating that our methods have significantly outperformed previous all baselines and achieved new state-of-the-art results. Moreover, qualitative analysis and the study on segmenting long-length sentences verify its effectiveness in modeling long-term dependencies.

\end{abstract}
	
\section{Introduction}
	
	Sequence segmentation, as an important task in natural language understanding (NLU), partitions a sentence into multiple segments. The first two rows of Table \ref{tab:Example Demo} show a case from a syntactic chunking dataset. The input sentence is a sequence of tokens and the output segments are multiple labeled phrases. These segments are nonoverlapping and fully cover the input sentence.
	
	In previous works, there are two dominant approaches to sequence segmentation. The most common is to regard it as a sequence labeling problem with resorting to IOB tagging scheme~\citep{huang2015bidirectional,akbik2018contextual,liu-etal-2019-gcdt}. This method is simple yet very effective, providing tons of state-of-the-art performances. For example, \newcite{huang2015bidirectional} present Bidirectional LSTM-CRF for named entity recognition (NER) and POS tagging, which adopts BiLSTM~\citep{hochreiter1997long} to read the input sentence and CRF~\citep{lafferty2001conditional} to decode the label sequence. An alternative method employs a transition-based system to incrementally segment and label an input token sequence~\citep{zhang2016transition,zhang2018simple}. For instance, \newcite{zhang2016transition} present a transition-based model for Chinese word segmentation that exploits not only character embedding but also token embedding. This type of method enjoys a number of attractive properties, including theoretically lower time complexity and capturing non-local features.
	
	\begin{table}[t]
		\centering
		
		\setlength{\tabcolsep}{1.1mm}{}
		\begin{tabular}{c|c}
			
			\hline
			\multirow{2}{*}{Sentence} & Tangible capital will be  about \\
			& \$ 115 million . \\
			
			\hline
			\multirow{2}{*}{Segments} & (Tangible capital, NP), (will be, VP), \\
			& (about \$ 115 million, NP), (., O)  \\
			\hline
			
			\multirow{2}{*}{IOB Tags} & B-NP I-NP B-VP I-VP \\
			
			& B-NP I-NP I-NP I-NP O  \\
			
			\hline
			& SHIFT SHIFT REDUCE-NP \\
			
			Transition &SHIFT SHIFT REDUCE-VP \\
			
			Actions	& SHIFT SHIFT SHIFT SHIFT \\
			
			& REDUCE-NP OUT \\
			
			\hline
		\end{tabular}
		
		\caption{The first two rows show an example extracted from CoNLL-2000 dataset~\citep{sang2000introduction}. The last two rows are two types of token-level labels commonly used to represent the segments.}
		\label{tab:Example Demo}
	\end{table}
	
	\begin{figure*}
		\centering
		\includegraphics[width=0.85\textwidth]{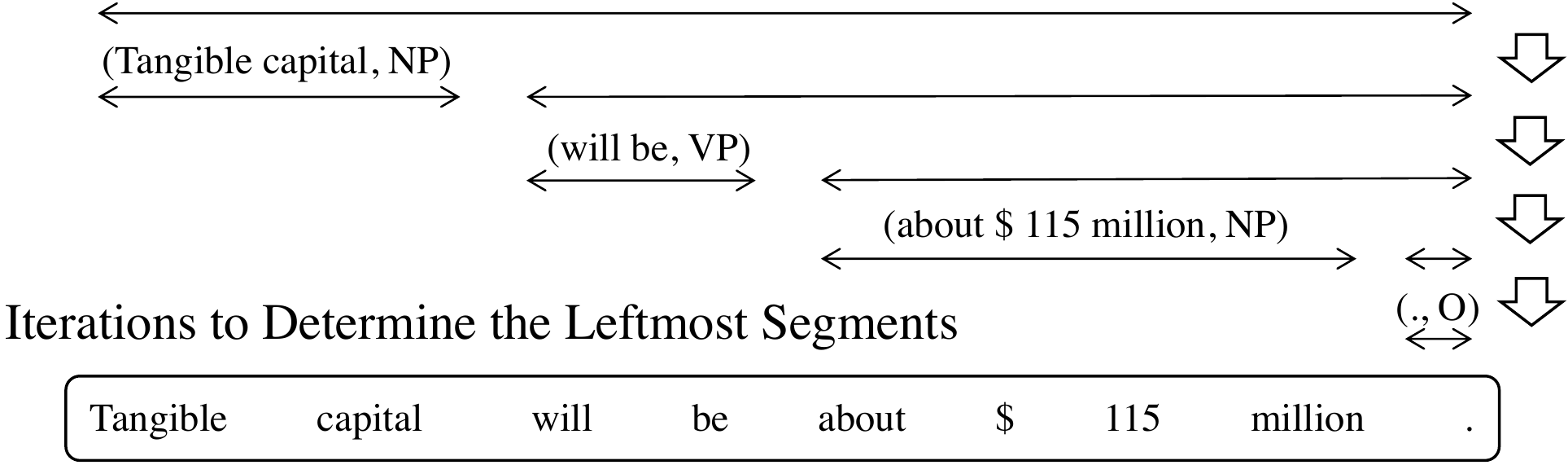}
		
		\caption{This case illustrates how our method segments a natural language sentence.}
		\label{fig:Approach}
	\end{figure*}
	
	The above two approaches are essentially at token level, where a single segment is represented by multiple token-level labels (e.g., transition actions). In spite of the adequacy, the labels used to model the relation among output segments are far more than the segments themselves. As demonstrated in Figure \ref{tab:Example Demo}, modeling the transition between the two segments, ``will be" and ``about \$ 115 million", consumes 6 IOB tags or 8 transition actions. This ill-posed design certainly limits the full potential of segmentation models to capture the long-term dependencies among segments.
	
	Previously, \citet{kong2015segmental} attempted to develop segment-level models, which define a joint probability distribution over the partition of an input sentence and the labeling of the segments. Such an approach circumvents using token-level labels. However, we find that, in experiments, it underperforms current token-level models. Moreover, its use of dynamic programming (DP) incurs quadratic running time, which is too slow for both training and inference (see Section \ref{sec:Running Time Analysis}).
	
	In this paper, we introduce a novel framework that incrementally segments a sentence at segment level. The segmentation process is iterative and incremental. At each iteration, the proposed framework determines the leftmost segment of the remaining unprocessed sequence. Under this scheme, we don't resort to token-level labels and enjoy linear time complexity. The implementation contains two stages. Firstly, we utilize LSTM-minus~\citep{wang2016graph,cross-huang-2016-span} technique to construct the representations for all the phrases. Secondly, we adopt LSTM to model the iterative segmentation process, which captures the strong correlation among segments. At every step, the input consists of the previous segment and the remaining unprocessed sequence, and the output is the leftmost segment.
	
	Figure \ref{fig:Approach} depicts how our framework segments the sentence in Table \ref{tab:Example Demo}. The output segments are obtained in an iterative and incremental manner. At each iteration, the leftmost segment of the remaining sequence is extracted and labeled. Compared with token-level models, we take much fewer steps to complete the segmentation process.
	
	Extensive experiments have been conducted on syntactic chunking and Chinese part-of-speech (POS) tagging across 3 datasets. The proposed framework has obtained new state-of-the-art performances on all of them. Besides, qualitative study and the results on segmenting long-length sentences confirm its effectiveness in capturing long-term dependencies.
	
	Our contributions are as follows:
	\begin{itemize}
		
		\item we present a novel framework that incrementally segments a natural language sentence at segment level. In comparison, previous approaches are mostly at token-level;
		
		\item we have notably outperformed previous baselines and established new state-of-the-art results on the 3 datasets of syntactic chunking and Chinese POS tagging. Experiments also show that our model is competitive with strong NER baselines;
		
		\item compared with prior methods, our model well captures the long-term dependencies among segments. This is strongly verified by qualitative study and the experiment on segmenting long-length sentences.
		
	\end{itemize}
	
	The source code of this work is available at https://github.com/LeePleased/LeftmostSeg.
	
\section{Architecture}

	We denote a $n$-length input sentence as $\mathbf{x} = [x_1, x_2, \cdots, x_n]$, where $x_i$ is a token (such as a word or a character). The output segments are represented as $\mathbf{y} = [y_1, y_2, \cdots, y_m]$, where $m$ is the amount of segments. Every segment $y_k$ is denoted as a triple $(i_k, j_k, l_k)$. $(i_k, j_k)$ is the span of the segment, corresponding to the phrase $\mathbf{x}_{i_k, j_k} = [\mathbf{x}_{i_k}, \mathbf{x}_{i_k + 1}, \cdots, \mathbf{x}_{j_k}]$. $l_k$ is from a predefined label space $\mathcal{L}$ and specifies the label of the segment. These segments are non-overlapping and fully cover the input sentence.
	
	\begin{figure*}
		\centering
		\includegraphics[width=0.8\textwidth]{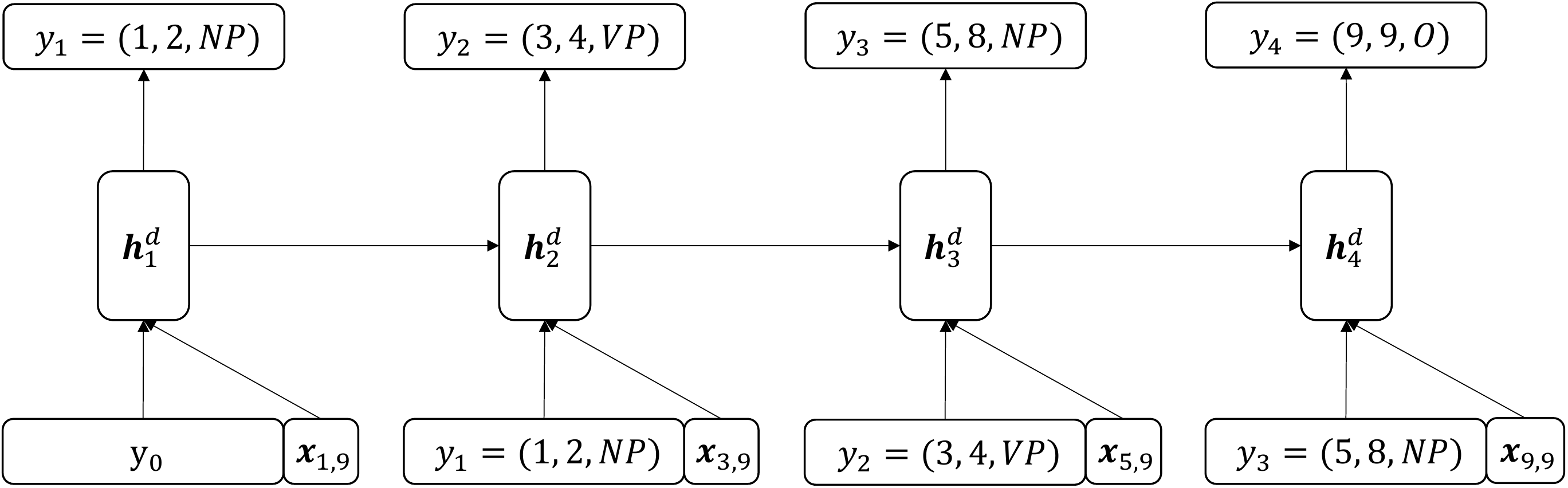}
		
		\caption{An example to show the incremental process to recognize leftmost segments.}
		\label{fig:Transduction}
	\end{figure*}
	
	The example in Table \ref{tab:Example Demo} is represented as
	\begin{equation}\nonumber
	\begin{aligned}
	\mathbf{x} =  [& \mathrm{Tangible}, \mathrm{capital}, \mathrm{will}, \mathrm{be}, \mathrm{about}, \mathrm{\$}, \mathrm{115}, \\
	& \mathrm{million}, \mathrm{.}] \\
	\mathbf{y} = [& (1, 2, \mathrm{NP}), (3, 4, \mathrm{VP}), (5, 8, \mathrm{NP}), (9, 9, \mathrm{O})]
	\end{aligned}.
	\end{equation}
	
\subsection{Phrase Representation Construction}
	
	Our goal here is to construct the representation for every phrase $\mathbf{x}_{i,j}$. Later, we will use the phrase representation to embed the segment $y_{k-1}$ and the unprocessed sequence $\mathbf{x}_{i_k, n}$.
	
	Firstly, each token $x_k$ is represented as 
	\begin{equation}
	\mathbf{e}_k = \mathbf{E}^{\text{t}}(x_k) \oplus \mathrm{CharCNN}(x_k),
	\end{equation}
	where $\mathbf{E}^\text{t}$ is a token embedding matrix and $\oplus$ is the column-wise vector concatenation. Following previous works~\citep{ma-hovy-2016-end,liu-etal-2019-gcdt}, we use $\mathrm{CharCNN}$ to extract the character-level representations.
	
	Secondly, we utilize bidirectional LSTMs $\overrightarrow{f}^{\text{e}}$ and $\overleftarrow{f}^{\text{e}}$ to compute the context-sensitive representation for each token $x_k$:
	\begin{equation}
	\left\{\begin{aligned}
	\overrightarrow{\mathbf{h}}^{\text{c}}_k & = \overrightarrow{f}^{\text{e}}(\overrightarrow{\mathbf{h}}^{\text{c}}_{k-1}, \mathbf{e}_k) \\ \overleftarrow{\mathbf{h}}^{\text{c}}_k & = \overleftarrow{f}^{\text{e}}(\overleftarrow{\mathbf{h}}^{\text{c}}_{k+1}, \mathbf{e}_k)  \\
	\mathbf{h}^{\text{c}}_k & = \overrightarrow{\mathbf{h}}^{\text{c}}_k \oplus \overleftarrow{\mathbf{h}}^{\text{c}}_k 
	\end{aligned}\right..
	\end{equation}
	
	Inspired by previous works~\citep{wang2016graph,gaddy-etal-2018-whats} in syntactic analysis, we integrate LSTM-minus features into phrase representations. Specifically, the representation $\mathbf{h}^\text{p}_{i, j}$ for a phrase $\mathbf{x}_{i,j}$ is computed as the concatenation of the difference of LSTM hidden states:
	\begin{equation} 
	\mathbf{h}^\text{p}_{i, j} = \mathbf{h}^{\text{c}}_j \oplus (\mathbf{h}^{\text{c}}_j - \mathbf{h}^{\text{c}}_i) \oplus \mathbf{h}^{\text{c}}_i.
	\end{equation}
	
	Inside algorithm~\citep{lari1990estimation} is another method to extracting phrase representation. Its advantage is to incorporate the potential hierarchical structure of natural language without using treebank annotation. For example, \citet{drozdov-etal-2019-unsupervised-latent} utilize inside algorithm to recursively compute the content representations. Despite the attractive property, its time complexity $\mathcal{O}(n^3)$  is too inefficient to practice use.
	
\subsection{Leftmost Segment Determination}
	
	Figure \ref{fig:Transduction} demonstrates how our method iteratively and incrementally segments the sentence in Table \ref{tab:Example Demo}. LSTM is used as the backbone to model the iterative process. At every step, the input consists of the prior segment and the unprocessed sequence, and the output is the predicted segment.

	Firstly, we embed the previous segment $y_{k-1}$ and the unprocessed sequence $\mathbf{x}_{i_{k},n}$. The previous segment is represented as
	\begin{equation}
		\mathbf{h}^\text{s}_{k-1} = \left\{\begin{array}{cc}
		\mathbf{h}^\text{p}_{i_{k - 1}, j_{k-1}} \oplus \mathbf{E}^\text{l}(l_{k-1}) & k > 1 \\
		\mathbf{v} & k = 1
		\end{array}\right.,
		\label{equ:Equation 7}
	\end{equation}
	where $\mathbf{E}^\text{l}$ is a label embedding matrix and $\mathbf{v}$ is a trainable vector. The unprocessed sequence is embedded as $\mathbf{h}^\text{p}_{i_k,n}$.
	
	At each iteration $k$, we use another LSTM $f^\text{d}$ to model the dependency among segments:
	\begin{equation}
	\mathbf{h}^\text{d}_k = f^\text{d}(\mathbf{h}^\text{d}_{k-1}, \mathbf{h}^\text{s}_{k-1} \oplus \mathbf{h}^\text{p}_{i_k,n}).
	\label{equ:Equation 8}
	\end{equation}
	During training, $i_k$ is known since the ground truth segments $\mathbf{y}$ is reachable. At evaluation time, we set $i_k = j_{k-1} + 1$.
	
	\begin{algorithm*}
		\caption{Inference Procedure}
		\label{algo:inference}
		
		\KwIn{The representations for all the phrases, $\mathbf{h}^\text{p}_{i,j}, 1 \le i \le j \le n$.}
		\KwOut{The sequence of predicted segments, $\hat{\mathbf{y}} = [\hat{y}_1, \hat{y}_2, \cdots, \hat{y}_{\hat{m}}]$.}
		
		Set a list $\mathbf{y}$ as $[]$ and set a counter as $k = 1$. \\
		Denote the remaining input token list as $\mathbf{x}$. \\
		Initialize the representation for previous segment as $\mathbf{v}$. \\
		Initialize the representation for unprocessed sequence as $\mathbf{h}^\text{p}_{1,n}$. \\
		
		\While{$\mathbf{x}$ is not empty}{
			Get LSTM hidden state $\mathbf{h}^\text{d}_k$ by using Equation \ref{equ:Equation 8}. \\
			Predict a segment $\hat{y}_k$ by using the Equations from \ref{equ:Equation 10} to \ref{equ:Equation 12}. \\
			Append the new segment $\hat{y}_k$ into list $\hat{\mathbf{y}}$. \\
			Reset the representation for previous segment as $\mathbf{h}^\text{p}_{\hat{i}_{k}, \hat{j}_{k}} \oplus \mathbf{E}^\text{l}(\hat{l}_{k})$. \\
			Reset the representation for unprocessed sequence as $\mathbf{h}^\text{p}_{\hat{j}_{k} + 1,n}$. \\
			Pop the tokens $[x_{\hat{i}_k}, x_{\hat{i}_k + 1}, \cdots, x_{\hat{j}_k}]$ from the remaining tokens $\mathbf{x}$. \\
			Increase the counter by 1: $k = k + 1$.
		}
		
		The amount of predicted segments: $\hat{m} = k$.
		
	\end{algorithm*}
	
	Then, we separately predict the span and the label of a segment. We define a set $\mathcal{S}_k$ containing all valid span candidates for prediction:
	\begin{equation}
		\mathcal{S}_k = \{(i_k, i_k), (i_k, i_k + 1), \cdots, (i_k, n)\}.
	\end{equation}
	The probability of a span $(i,j) \in \mathcal{S}_k$ is
	\begin{equation}
	Q^\text{s}_{k,i,j} \propto \exp \big((\mathbf{h}^\text{d}_k)^{\top}\mathbf{W}^\text{s}\mathbf{h}^\text{p}_{i, j} \big).
	\label{equ:Equation 10}
	\end{equation}
	The probability of a label $l \in \mathcal{L}$ for a span $(i, j)$ is
	\begin{equation}
		Q^\text{l}_{k,i,j,l} \propto \exp\big(\mathbf{E}^\text{l}(l)^{\top} \mathbf{W}^\text{l} (\mathbf{h}^\text{p}_{i,j} \oplus \mathbf{h}^\text{d}_k ) \big).
		\label{equ:Equation 11}
	\end{equation} 
	The matrices $\mathbf{W}^\text{s}$ and $\mathbf{W}^\text{l}$ are learnable.
	
	Finally, the leftmost segment is obtained as
	\begin{equation}
	\left\{\begin{aligned}
	(\hat{i}_k, \hat{j}_k) & = \mathop{\arg\max}_{(i,j) \in \mathcal{S}_k} Q^\text{s}_{k,i,j} \\
	\hat{l}_k & = \mathop{\arg\max}_{l \in \mathcal{L}} Q^\text{l}_{k, i, j, l} \\
	\hat{y}_k & = (\hat{i}_k, \hat{j}_k, \hat{l}_k)
	\end{aligned}\right..
	\label{equ:Equation 12}
	\end{equation}
	
	The iterative process ends when the remaining sequence $\mathbf{x}_{j_k + 1, n}$ is empty (i.e., $j_k = n$).
	
\subsection{Training and Inference}
	
	During training, we use teacher forcing where every segment $y_k$ is predicted using its previous ground-truth segments $[y_1, y_2, \cdots, y_{k-1}]$. A hybrid loss is induced as
	\begin{equation}
	\mathcal{J} = - \sum_{y_k \in \mathbf{y}} (\log Q^\text{s}_{k,i_k,j_k} + \log Q^\text{l}_{k,i_k,j_k,l_k} ).
	\label{equ:Equation 13}
	\end{equation}
	
	At test time, every segment $\hat{y}_k$ is inferred in terms of the previous predicted segments $[\hat{y}_{1}, \hat{y}_2, \cdots, \hat{y}_{k-1}]$. Algorithm \ref{algo:inference} demonstrates how our proposed framework makes inference. Note that Algorithm \ref{algo:inference} uses greedy search to get $\hat{y}$ because it is both fast in speed and effective in accuracy in our experiments, although beam search may be better in accuracy.
	
\section{Experiments}
	
	\begin{table*}
		\centering
		
		\setlength{\tabcolsep}{7.1mm}{}
		\begin{tabular}{c|c|c}
				
				\hline
				\multicolumn{2}{c|}{Approach} & CoNLL-2000 \\
				
				\hline
				\multicolumn{2}{c|}{Segmental RNN~\citep{kong2015segmental}} & $95.08$ \\
				\multicolumn{2}{c|}{Bi-LSTM + CRF~\citep{huang2015bidirectional}} & $94.46$ \\
				\multicolumn{2}{c|}{Char-IntNet-5~\citep{xin-etal-2018-learning}} & $95.29$ \\
				\multicolumn{2}{c|}{GCDT~\citep{liu-etal-2019-gcdt}} & $95.17$ \\
				
				\cdashline{1-3}
				\multicolumn{2}{c|}{Flair Embedding~\citep{akbik2018contextual}} &$96.72$  \\
				\multicolumn{2}{c|}{Cross-view Training~\citep{clark-etal-2018-semi}}  & $97.00$ \\
				\multicolumn{2}{c|}{GCDT w/ BERT~\citep{liu-etal-2019-gcdt}} & $96.81$ \\
				
				\hline 
				\multirow{2}{*}{This Work} & Our Model & $\mathbf{96.13}$ \\
				& Our Model w/ BERT & $\mathbf{97.05}$ \\
				
				\hline
				
		\end{tabular}
		
		\caption{The performances of the baselines and our models on CoNLL-2000 dataset.}
		\label{tab:Experiment on Syntactic Chunking}
	\end{table*}

	\begin{table*}
		\centering
		
		\setlength{\tabcolsep}{7.1mm}{}
		\begin{tabular}{c|c|cc}		
			
			\hline
			\multicolumn{2}{c|}{Approach} & PTB9 & UD1 \\
			
			\hline
			\multicolumn{2}{c|}{Segmental RNN~\citep{kong2015segmental}} & $92.16$ & $90.01$ \\
			\multicolumn{2}{c|}{Bi-RNN + CRF (single)~\citep{shao-etal-2017-character}} & $91.89$ & $89.41$ \\
			\multicolumn{2}{c|}{Bi-RNN + CRF (ensemble)~\citep{shao-etal-2017-character}} & $92.34$ & $89.75$ \\
			\multicolumn{2}{c|}{Lattice LSTM~\citep{zhang-yang-2018-chinese}} & $92.13$ & $90.09$ \\
			\multicolumn{2}{c|}{Glyce + Lattice LSTM~\citep{meng2019glyce}} & $92.38$ & $90.87$ \\
			
			\cdashline{1-4}
			\multicolumn{2}{c|}{BERT~\citep{devlin-etal-2019-bert}} & $92.29$ & $94.79$  \\
			\multicolumn{2}{c|}{Glyce + BERT~\citep{meng2019glyce}} & $93.15$ & $96.14$ \\
			
			\hline 
			\multirow{2}{*}{This Work} & Our Model & $\mathbf{92.56}$ & $\mathbf{91.65}$ \\
			& Our Model w/ BERT & $\mathbf{93.38}$ & $\mathbf{96.43}$ \\
			
			\hline
				
		\end{tabular}
		\caption{The results on the two datasets of Chinese POS tagging.}  
		\label{tab:Experiment on Chinese POS tagging}
	\end{table*}
	
	Extensive experiments have been conducted on syntactic chunking and Chinese POS tagging across 3 datasets. Firstly, our models have obtained new state-of-the-art performances on all the datasets. Then, we have investigated ablation studies to understand the importance of each component. Lastly, case study and the results on segmenting long-length sentences confirm the effectiveness of the proposed framework in capturing the long-term dependencies among segments.
	
\subsection{Settings}
	
	Syntactic chunking segments a word sequence into multiple labeled groups of words. We use CoNLL-2000 dataset~\citep{sang2000introduction}, which defines 11 syntactic chunk types (NP, VP, PP, etc.). Standard data includes a training set and a test set. Following~\citet{xin-etal-2018-learning}, we randomly sample 1000 sentences from the training set as the development set. Chinese POS tagging converts a Chinese character sequence into a token sequence and associates every word with a POS tag. We use Penn Chinese Treebank 9.0 (CTB9)~\citep{xue2005penn} and Universal Dependencies 1.4 (UD1)~\citep{nivre2016universal}. CTB9 contains the source text in various genres, covering its previous versions (e.g., CTB6). We use the Chinese section of UD1. We follow the same format and partition of the two datasets as~\citet{shao-etal-2017-character}.
	
	We use the same neural network configuration for all 3 datasets. The dimensions of token embedding and label embedding are respectively set as $300$ and $50$. The hidden unit sizes for the encoder and decoder are $256$ and $512$, respectively. The layers of two LSTMs are both $2$. L2 regularization is set as $1 \times 10^{-6}$ and dropout ratio is set as $0.4$ for reducing overfit. The above setting is obtained by grid search. We adopt Adam~\citep{kingma2014adam} as the optimization algorithm and adopt the suggested hyper-parameters. For CoNLL-2000 dataset, the cased, 300d Glove~\citep{pennington2014glove} is used to initialize token embedding. $\mathrm{CharCNN}$ is not used in Chinese tasks. The batch size is set as 16. All our models in experiments are running on NVIDIA Tesla P100.
	
	At test time, following previous literature, we convert the prediction of our model into IOB format and use the standard conlleval script\footnote{https://www.clips.uantwerpen.be/conll2000/chunking/co\\nlleval.txt.} to get the F1 score.  We select the model that works the best on development set, and then evaluate it on test set. In all the experiments, the improvements of our models over the baselines are statistically significant with $p < 0.05$ under t-test.
	
\subsection{Results on Syntactic Chunking} 
\label{sec:Results on Syntactic Chunking}

	\begin{table*}
		\centering

		\begin{tabular}{c|ccc}
			
			\hline
			Approach & CoNLL-2000 & PTB9 & UD1 \\
			
			\hline
			Our Model & $\mathbf{96.13}$ & $\mathbf{92.56}$ & $\mathbf{91.65}$ \\
			
			\hline
			w/o $\mathrm{CharCNN}$ & $95.81$ & - & - \\
			w/o LSTM-minus, w/ Inside Algorithm & $\mathbf{96.35}$ & $\mathbf{92.71}$ & $\mathbf{91.87}$ \\
			
			\cdashline{1-4}
			w/o LSTM Decoder $f^d$, w/ MLP & $94.91$ & $91.28$ & $90.05$ \\
			
			w/o Phrase Representation $\mathbf{h}^p_{i_{k - 1}, j_{k-1}}$ in Equation \ref{equ:Equation 7} & $95.78$ & $92.09$ & $91.27$ \\
			
			w/o Label Representation $\mathbf{E}^l(l_{k-1})$ in Equation \ref{equ:Equation 7} & $95.82$ & $91.92$ & $91.04$ \\
			
			\cdashline{1-4}
			w/o Greedy Search, w/ Beam Search & $\mathbf{96.22}$ & $\mathbf{92.77}$ & $\mathbf{91.72}$  \\
			
			\hline
		\end{tabular}
		
		\caption{The results of ablation experiments on all three datasets.}  
		\label{tab:Abalation Studies}
	\end{table*}
	
	Our models are compared with two groups of baselines. One of them is trained without any external resources besides the training data:
	\begin{description}
		
		\item[Segmental RNN] It's a segment-level model that defines a joint probability distribution over the partition of an input sequence and the labeling of the segments;
		
		\item[Bi-LSTM + CRF] It utilizes bidirectional LSTM to read the input sentence and CRF to decode the label sequence;
		
		\item[Char-IntNet-5] It is a funnel-shaped CNN model with no down-sampling which learns a better internal structure for tokens;
		
		\item[GCDT] It deepens the state transition path at each position in a sentence and assigns each token with a global representation learned from the entire sentence.
		
	\end{description}
	
	The other uses extra unlabeled corpora or fine-tunes on a pre-trained language model:
	\begin{description}
		
		\item[Flair Embedding] It firstly pre-trains a character-level language model on a large corpus, and then uses a sequence labeling model (e.g., Bi-LSTM + CRF) to fine-tune on it;
		
		\item[Cross-view Training] It designs a LSTM based sentence encoder to facilitate semi-supervised learning. The model can benefit from massive unlabeled corpora;
		
		\item[GCDT w/ BERT] It adopts BERT as additional token embeddings to improve GCDT. 
		
	\end{description}
	
	We adopt most of the results of baselines as reported in~\citet{huang2015bidirectional,akbik2018contextual,xin-etal-2018-learning}. Since the evaluation method of GCDT is not standard (see the experiment setup in \citet{luo2020hierarchical}), we correct its source code\footnote{https://github.com/Adaxry/GCDT.} to retest the performance. The result for Segmental RNN is from our re-implementation.
	
	Table \ref{tab:Experiment on Syntactic Chunking} demonstrates that we have notably outperformed previous methods and achieved new state-of-the-art results on CoNLL-2000 dataset. When not using external resource, we obtain the F1 score of $96.13$, which outperforms Segmental RNN by $1.10\%$, Bi-LSTM + CRF by $1.77\%$, Char-IntNet-5 by $0.88\%$, and GCDT by $1.01\%$. Note that Segmental RNN, a segment-level model, underperforms Char-IntNet-5, a token-level model, by $0.22\%$. To make a fair comparison with the baselines using additional unlabeled corpora, we also use BERT~\citep{devlin-etal-2019-bert}, a powerful pretrained language model, to replace our token embedding. In this way, we achieve the F1 score of $97.05$, which outnumbers Flair Embedding by $0.34\%$, Cross-view Training by $0.05\%$, and GCDT w/ BERT by $0.25\%$. All these results verify the effectiveness of our framework.
	
\subsection{Results on Chinese POS Tagging}
\label{sec:Results on Chinese POS Tagging}
	
	We categorize the baselines into two types. The methods without using external resources: 
	\begin{description}
		
		\item[Segmental RNN] Also described in Section \ref{sec:Results on Syntactic Chunking};
		
		\item[Bi-RNN + CRF] It utilizes bidirectional RNN and CRF to model joint word segmentation and POS tagging. Ensemble learning is also used to improve the results;
		
		\item[Lattice LSTM] It's a lattice-structured LSTM that encodes Chinese characters as well as all potential tokens that match a lexicon;
		
		\item[Glyce + Lattice LSTM] It incorporates Chinese glyph information into Lattice LSTM.
		
	\end{description}
	
	Others using a pre-trained language model:
	\begin{description}
		
		\item[BERT] It is a language model pre-training on a large corpus. It uses the representations from the last layer to predict IOB tags;
		
		\item[Glyce + BERT] It integrates Chinese glyph information into BERT Tagging model.
		
	\end{description}

	We take most of the performances of baselines from \citet{meng2019glyce}. The results for Segmental RNN are from our re-implementation.
	
	Table \ref{tab:Experiment on Chinese POS tagging} shows that our models have achieved state-of-the-art results on the two datasets, PTB9 and UD1. When BERT is not used, we have obtained the F1 scores of $92.56$ and $91.65$, which outperform Glyce + Lattice LSTM by $0.19\%$ and $0.86\%$ and Bi-RNN + CRF (ensemble) by $0.24\%$ and $2.12\%$. Note that Segmental RNN, a segment-level model, underperforms Lattice LSTM, a token-level model, by $0.09\%$ on UD1. When using BERT, even without incorporating Chinese glyph information, we still obtain the F1 scores of $93.38$ and $96.43$, which outperform Glyce + BERT by $0.25\%$ and $0.30\%$. These results further confirm the effectiveness of our proposed framework.
	
\subsection{Ablation Studies}
	
	\begin{table*}
		\centering

		\begin{tabular}{c|cccc|c}
			
			\hline
			Approach &  1-22 (711) & 23-44 (1030) & 45-66 (248) & 67-88 (23) & Overall \\
			
			\hline
			Bi-LSTM + CRF & $94.01$ & $94.76$ & $92.98$ & $87.09$ & 94.23 \\
			
			GCDT & $94.95$ & $95.52$ & $93.77$ & $87.14$ & 95.17 \\
			
			\hline
			Our Model & $\mathbf{96.01}$ & $\mathbf{96.55}$ & $\mathbf{94.82}$ & $\mathbf{91.07}$ & $\mathbf{96.13}$ \\
			
			\hline
		\end{tabular}
		
		\caption{The F1 scores for the sentences of different length ranges.}  
		\label{tab:Segmenting Long-length Sentences}
	\end{table*}
	
	\begin{table*}
		\centering
		
		\setlength{\tabcolsep}{1.1mm}{}
		\begin{tabular}{c|c|c}
			
			\hline
			\multicolumn{2}{c|}{\multirow{2}{*}{Input Sentence}} & Other antibodies sparked by the preparation are of a sort \\
			\multicolumn{2}{c|}{} & rarely present in large quantities in infected or ill individuals \\
			
			\hline
			\multirow{6}{*}{Output Segments} & \multirow{3}{*}{GCDT} & (Other antibodies, NP) (sparked, VP) (by, PP) (the preparation, NP) \\
			
			& & (are, VP) (of, PP) (a sort, NP) \textbf{(rarely present, VP)} (in, PP)\\
			
			& & (large quantities, NP) (in, PP) (infected or ill individuals, NP) \\
			
			\cline{2-3}
			& \multirow{3}{*}{Our Model} & (Other antibodies, NP) (sparked, VP) (by, PP) (the preparation, NP) \\
			
			& & (are, VP) (of, PP) (a sort, NP) (rarely present, ADJP) (in, PP)\\
			
			& & (large quantities, NP) (in, PP) (infected or ill individuals, NP) \\
			
			\hline
			
		\end{tabular}
		\caption{The case is from CoNLL-2000 dataset. The predicted segments of our model is correct.}  
		\label{tab:Case Study}
	\end{table*}
	
	As shown in Table \ref{tab:Abalation Studies}, we conduct ablation studies to explore the impact of every component.
	
	\paragraph{Effect of Representation Learning.} Following prior works~\citep{ma-hovy-2016-end,liu-etal-2019-gcdt}, we employ $\mathrm{CharCNN}$ to incorporate character information into word representations. By removing it, the F1 score on CoNLL-2000 decreases by $0.33\%$. Inside algorithm is another technique to construct phrase representations. After using it to replace LSTM-minus, the results on the three datasets are slightly improved by $0.23\%$, $0.16\%$, and $0.24\%$. Our implementation of inside algorithm is the same as described in \citet{drozdov-etal-2019-unsupervised-latent}. Despite the slight improvements, its time complexity $\mathcal{O}(n^3)$ is too slow for both training and inference. Empirically, we find that the running time of inside algorithm is about $7$ times slower than that of LSTM-minus. 
	
	\paragraph{Effect  of Modeling the Dependencies Among Segments.} LSTM decoder $f^d$ models the long-term dependencies among segments. The prediction of every segment $y_k$ is conditional on prior segments $y_{k'}, 1 \le k' < k$. By replacing it with multilayer perceptron (MLP), the performances fall by $1.29\%$, $1.40\%$, and $1.78\%$ on the three datasets. We use both phrase representation $\mathbf{h}^p_{i_{k - 1}, j_{k-1}}$ and label representation $\mathbf{E}^l(l_{k-1})$ to embed the previous segment $y_{k-1}$. After removing the phrase representations, the F1 scores decrease by $0.37\%$, $0.51\%$, and $0.42\%$ on the three datasets. By removing the label representations, the results also drop by $0.32\%$, $0.70\%$, and $0.67\%$.
	
	\paragraph{Effect of Inference Algorithm} Beam search is a widely used technique in language generation tasks, like machine translation~\citep{bahdanau2014neural,NIPS2017Vaswani,li2020rewriter} and data-to-text generation~\citep{shen-etal-2020-neural,li-etal-2020-slot,li2020interpretable}. We have attempted to use beam search (with the beam size being $5$), instead of greedy search, for inference. By doing so, the F1 scores of our models increased by $0.09\%$ on CoNLL-2000, $0.23\%$ on PTB9, and $0.08\%$ on UD1. While obtaining slightly better performances, the time costs for inference become intolerably high (increase about $10$ times). Therefore, we adopt greedy search as the default inference algorithm.

\subsection{Segmenting Long-length Sentences}
\label{sec:Segmenting Long-length Sentences}

	\begin{table*}
		\centering

		\begin{tabular}{c|c|cc}
			
			\hline
			Approach &  Time Complexity & Training Time & Evaluation Time \\
			
			\hline
			Segmental RNN & $\mathcal{O}(n^2|\mathcal{L}|^2)$ &16m17s & 3m01s \\
			
			Bi-LSTM + CRF & $\mathcal{O}(n|\mathcal{L}|^2)$ & 3m28s & 0m26s \\
			
			\hline
			Our Model & $\mathcal{O}(n|\mathcal{L}|)$ & 2m36s & 0m20s \\
			
			\hline
		\end{tabular}
		
		\caption{Running time comparisons on CoNLL-2000 dataset.}  
		\label{tab:Running Time Comparisons}
	\end{table*}

	\begin{table*}
		\centering

		\begin{tabular}{c|c|c}		
			
			\hline
			Approach & CoNLL-2003 & OntoNotes 5.0 \\
			
			\hline
			BiLSTM-CNN-CRF~\citep{chen2019grn} & $91.21$ & $87.05$ \\
			
			GRN~\citep{chen2019grn} & $91.44$ & $87.67$ \\
			
			HCR~\citep{luo2020hierarchical}  & $\mathbf{91.96}$ & $\mathbf{87.98}$ \\
			
			\hline
			Our Model & $91.42$ & $87.74$ \\
			
			\hline
		\end{tabular}
		\caption{The results on two NER datasets.}  
		\label{tab:Experiment on NER}
	\end{table*}
	
	Compared with token-level methods, our framework better captures the long-term dependencies among segments. Therefore, our model should be more accurate in segmenting long-length sentences. To verify this, we test the baselines and our model on the sentences of different lengths.
	
	The study is conducted on CoNLL-2000 dataset. Bi-LSTM + CRF and GCDT are very strong baselines and have open source implementations. We use toolkit NCRFPP\footnote{https://github.com/jiesutd/NCRFpp.} to reproduce the performances of Bi-LSTM + CRF. We use the reproduction in Section \ref{sec:Results on Syntactic Chunking} as the results of GCDT. Table \ref{tab:Segmenting Long-length Sentences} shows the experiment results. Each column name denotes the sentence length range and the case number. Our model notably outperforms prior methods in terms of long sentence length ranges. For length range 45-66, our model obtains the F1 score of $94.82$, which outperforms Bi-LSTM + CRF by $1.98\%$ and GCDT by $1.12\%$. For length range 67-88, our model outperforms Bi-LSTM + CRF by $4.57\%$ and GCDT by $4.51\%$.
	
\subsection{Case Study}
	
	In Table \ref{tab:Case Study}, we present an example extracted from the test set of CoNLL-2000. Given an input sentence, we show the prediction from a strong baseline, GCDT, and our model. The output segments from our model are consistent with the ground truth segments. The segment in bold is the one incorrectly produced by GCDT.
	
	Understanding the structure of this sentence is very hard because the main constituents ``Other antibodies", ``are", and ``rarely present" locate far apart. By removing the two middle phrases ``sparked by the preparation" and ``of a sort", the original sentence can be simplified to ``Other antibodies are rarely present in large quantities in infected or ill individuals", which is very clear. Therefore, correctly predicting the segment, (rarely present, ADJP), implies that our model well captures the long-term dependencies among the segments. For GCDT, a token-level segmentation model, it mistakes ``rarely present" for the verb phrase of ``a sort". This may result from the following two causes: 
	\begin{itemize}
		
		\item The adjacent segments labeled with (NP, VP) frequently appear in training data;
		
		\item the token ``present" acting as a verb is much more common than as a adjective.
		
	\end{itemize}
	Both potentials indicate that token-level models can't capture the long-term dependencies well.
	
\subsection{Running Time Analysis}
\label{sec:Running Time Analysis}
	
	Table \ref{tab:Running Time Comparisons} demonstrates the running time comparison among different methods. The last two columns are respectively the running times for training (one epoch) and evaluation. We set the batch size as 16 and run all the models on 1 GPU. From the table, we can draw the following two conclusions. Firstly, Segmental RNN, a segment-level model, is very slow for both training and inference due to the high time complexity. For instance, its training and testing are respectively $6.26$ and $9.05$ times slower than ours. Secondly, our framework is very efficient. For example, training our model for one epoch is $1.33$ times faster than training Bi-LSTM + CRF, a token-level model.
	
\subsection{Results on NER}

	While our model is tailored for sequence segmentation tasks, we have also tested its performances on two widely used NER datasets, CoNLL-2003~\citep{sang2003introduction} and OntoNotes 5.0~\citep{pradhan2013towards}. Note that, in NER, the label correlations among adjacent segments are very weak. This seems bad for our model.
	
	Table \ref{tab:Experiment on NER} diagrams the comparison of our model and strong NER baselines. The results of GRN and HCR are copied from \citet{chen2019grn,luo2020hierarchical}. For BiLSTM-CNN-CRF, its scores on CoNLL-2003 and OntoNotes 5.0 are respectively from \citet{chen2019grn} and our re-implementation. From the table, we can see that our model is competitive with prior methods. For example, our model underperforms HCR by only $0.27\%$ on OntoNotes 5.0. In particular, our F1 scores are notably higher than those of BiLSTM-CNN-CRF by $0.23\%$ and $0.79\%$ on the two datasets. This experiment shows that our model is applicable to general sequence labeling tasks.

\section{Related Work}
	
	There are two mainstream approaches to sequence segmentation. One of them treat sequence segmentation as a sequence labeling problem by using IOB tagging scheme~\citep{huang2015bidirectional,xin-etal-2018-learning,clark-etal-2018-semi,akbik2018contextual,liu-etal-2019-gcdt,li-etal-2020-handling}. Each token in a sentence is labeled as B-tag if it’s the beginning of a segment, I-tag if it is inside but not the first one within the segment, or O otherwise. This method is extensively studied by prior works and provides tons of state-of-the-art results.  \newcite{xin-etal-2018-learning} introduce a funnel-shaped convolutional architecture that learns a better internal structure for the tokens. \newcite{akbik2018contextual} propose an efficient character-level framework that uses pretrained character embedding. \newcite{clark-etal-2018-semi} adopt semi-supervised learning to train LSTM encoder using both labeled and unlabeled corpora. Despite the effectiveness, these models are at token level, relying on multiple token-level labels to represent a single segment. This limits their full potential to capture the long-term dependencies among segments. 
	
	The other uses transition-based systems as the backbone to incrementally segment and label an input sequence~\citep{qian2015transition,zhang2016transition,zhang2018simple}. For example, \newcite{qian2015transition} design special transition actions to jointly segment, tag, and normalize a sentence. These models have many advantages, such as theoretically lower time complexity and capturing non-local features. However, they are still token-level models, which predict transition actions to shift a token from the buffer to the stack or assign a label to a span.
	
	Recently, there is a surge of interest in developing span-based models, such as Segmental RNN~\citep{sarawagi2004semi,kong2015segmental} and LUA~\citep{li2020segmenting}. These methods circumvent using token-level labels and directly label the phrases in a sentence. Span-based models also enjoy great popularity in language modeling~\citep{li2020span}, NER~\citep{yu-etal-2020-named,li2021empirical}, and constituent parsing~\citep{cross-huang-2016-span,stern-etal-2017-minimal}. However, its underperforms current token-level models (see Section \ref{sec:Results on Syntactic Chunking} and Section \ref{sec:Results on Chinese POS Tagging}) and is very slow in terms of running time (see Section \ref{sec:Running Time Analysis}).

\section{Conclusion}

	In this work, we present a novel framework to sequence segmentation that segments a sentence at segment level. The segmentation process is iterative and incremental. For every step, it determines the leftmost segment of the remaining sequence. Implementations involve LSTM-minus to extract phrase representations and RNN to model the iterations of leftmost segment determination. Extensive experiments have been conducted on syntactic chunking and Chinese POS tagging across 3 datasets. We have achieved new state-of-the-art performances on all of them. Case study and the results on segmenting long-length sentences both verify the effectiveness of our framework in modeling long-term dependencies.
	
\section*{Acknowledgments}
	
	 This work was done when the first author did internship at Ant Group. We thank anonymous reviewers for their kind and constructive suggestions.
	
\bibliography{anthology,custom}
\bibliographystyle{acl_natbib}
	
\end{document}